# Challenge report: Recognizing Families In the Wild Data Challenge

*Zhipeng Luo, Zhiguang Zhang, Zhenyu Xu, Lixuan Che*
DeepBlue Technology
DeepBlueAI
Beijing, China
{ luozp, zhangzhg, xuzy, chelx}@ deepblueai.com

*Abstract*—This paper is a brief report to our submission to the Recognizing Families In the Wild Data Challenge (4th Edition), in conjunction with FG 2020 Forum. Automatic kinship recognition has attracted many researchers' attention for its full application, but it is still a very challenging task because of the limited information that can be used to determine whether a pair of faces are blood relatives or not. In this paper, we studied previous methods and proposed our method. We try many methods, like deep metric learning-based, to extract deep embedding feature for every image, then determine if they are blood relatives by Euclidean distance or method based on classes. Finally, we find some tricks like sampling more negative samples and high resolution that can help get better performance. Moreover, we proposed a symmetric network with a binary classification based method to get our best score in all tasks.

## I. INTRODUCTION

The appearance of a person is mostly genetically determined, hence there are apparent similarities in the looks of a family. Families In the Wild (FIW) is a large dataset[1-4] that aims to identify the family members in the wild according to the similarities in their appearance. Kinship recognition is beneficial for looking for the missing child, lost Alzheimer's patient, and so on. The tasks of "Recognizing Families In the Wild Data Challenge (4th Edition) in conjunction with FG 2020" can be roughly divided into the following tasks:

1) Kinship Verification (one-to-one): Kinship verification aims to determine whether a pair of facial images are blood relatives of a certain type.

2) Tri-subject Verification (one-to-two): Tri-Subject Verification focuses on a slightly different view of kinship verification– the goal is to decide whether a child is related to a pair of parents.

3) Search and Retrieval (many-to-many): Large-Scale Search and Retrieval of family members of missing children. The goal is to find family members of the search subjects (i.e., the probes) in a search pool (i.e., the gallery).

In this paper, we introduce some methods we try and the best method that gets our best score in all tasks, which is a binary classification based method with a symmetric network.

## II. METHOD

In this report, we first refer to the 1st place solution of "Recognizing Faces in the Wild" in kaggle[5] as our baseline and make some necessary adaptations for solving the targeted challenge better.

### A. Datasets

For task 1, we merge the training and validation set as our training dataset, include 763 families. Fig.1 shows an example of the given samples, where each family folder contains multiple family members, with F0300 being the family id, and MID1 being an individual in the family with the id of 1. MID1 has two pictures available. The file mid.csv contains information on the names of family members and a relationship matrix, and its family-level labels, as shown in Table I. The relationship matrix is built upon nine types of labels, as shown in Table II. Datasets for task 2 and task 3 have similar size and identical structure with that of task 1.

```
Train-faces
F0300
 └─MID1
 │  └─P0319_face0.jpg
 │  └─P0319_face1.jpg
 └─MID2
 │  └─P0326_face0.jpg
 └─mid.csv
 └─unrelated_and_nonfaces
    └─P0398_face0.jpg
    └─P0399_face1.jpg
```

Fig. 1. The folder structure of FIW dataset

TABLE I.   RELATIONSHIP MATRIX

| MID | 1 | 2 | 3 | Name | Gender |
|-----|---|---|---|------|--------|
| 1 | 0 | 4 | 5 | Name1 | Female |
| 2 | 1 | 0 | 1 | Name2 | Female |
| 3 | 5 | 4 | 0 | Name3 | Female |

TABLE II.   RELATIONSHIP ID (RID) AND KEY VALUES

| RID | Label |
|-----|-------|
| 1 | Child |
| 2 | Sibling |
| 3 | Grandchild |
| 4 | Parent |
| 5 | Spouse |
| 6 | Grandparent |
| 7 | Great Grandchild |
| 8 | Great Grandparent |
| 9 | TBD |
| 0 | NA |

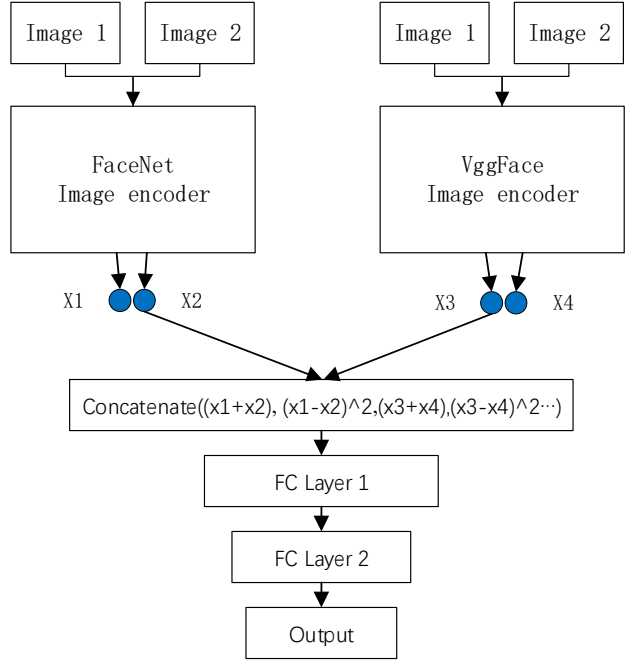

Fig.2. The pipeline used in this paper, First extract features for each image using FaceNet and VggFace, then concatenate the transformed features and feed them into fully connected layers to obtain probabilities of kinship.

In the nine types of relationships, only 1,2,3,4,6,7,8 denote blood relatives and hence can be considered as positive pairs, while others are negative ones.

*B. Model*

As shown in Fig.2, we design a network based on the Siamese structure. Specifically, we first extract features for each image using FaceNet[6] and VggFace[7]. This way, images are encoded into fixed-length vectors. We then concatenate the transformed features and feed them into fully connected layers. Finally, we obtain a binary or multi-class probabilities of kinship.

To get better performance, we use a pre-trained FaceNet encoder on MS-Celeb-1M[8] dataset, a benchmark for Large-Scale Face Recognition task. Besides, we also use a pre-trained VggFace encoder trained on vggface2[7] datasets, a dataset of 3.3 million face images and 9000+ identities in a wide range of different ethnicities, accents, professions, and ages, for face recognition task too.

In the concatenate stage, we apply some transforms on the features to promote discrimitivity, such as (x1^2-x2^2), (x1-x2)^2, (x1x2),(x1+x2), (x1-x2), (sqrt(x1) + sqrt(x2)), (sqrt(x1) - sqrt(x2)), (x3^2-x4^2), etc.

*C. Loss function*

We use two different methods to predict the labels for a given input image pair.

Firstly, we take the task as a binary classification problem, using sigmoid to predict a probability indicating whether the given images are blood relative. For this method, we use Binary CrossEntropy Loss(BCE)[9] and Focal loss[10] as the loss function independently.

In BCE, the ground truth includes label y=1 indicate there are blood relatives between input images, and label y=0 otherwise:

$$L_{bce} = -[y \log(p) + (1-y) \log(1-p)] \quad (1)$$

Focal loss is a loss function to solve the problem of imbalance of positive and negative samples in the task of object detection. It is reconstructed from BCE and can be used for binary classification problems as well. In this paper, we set hyper-parameters to α=0.25, γ=2:

$$L_{fl} = \begin{cases} -\alpha(1-p)^\gamma \log(p) & y=1 \\ -(1-\alpha)p^\gamma \log(1-p) & y=0 \end{cases} \quad (2)$$

Second, we take the task as a multi-class classification problem. We divide the relationships that have blood relative into three categories based on whether they are in the same generation or not and add one additional class to indicate the case where no blood relative, as shown in Table III.

TABLE III.    MULTI-CLASS LABEL

| Type in the relationship matrix | New class |
|---|---|
| 1,4 | 1 |
| 2 | 2 |
| 3,6,7,8 | 3 |
| others | 0 |

In this method, we use softmax to predict the probabilities belonging to each class, where $z_i$ is the output of the network:

$$P(y = j \mid x) = \frac{e^{z_i}}{\sum_{k=1}^{K} e^{z_k}} \quad (3)$$

TABLE IV. RESULT OF BASELINE AND DIFFERENT IMAGE INPUTS AND LOSS FUNCTIONS

| Exp_id | Backbone | RGB or Gray | Loss function | Val_Acc | Test score |
|---|---|---|---|---|---|
| 1 | SeNet50 | RGB | BCE | 0.77438 | 0.7085 |
| 2 | ResNet50 | RGB | BCE | 0.77000 | - |
| 3 | SeNet50 | RGB | Focal loss | 0.75125 | - |
| 4 | ResNet50 | RGB | Focal loss | 0.74250 | - |
| 5 | SeNet50 | Gray | BCE | 0.76000 | - |
| 6 | ResNet50 | Gray | BCE | 0.76062 | - |
| 7 | SeNet50 | Gray | Focal loss | 0.74125 | - |
| 8 | ResNet50 | Gray | Focal loss | 0.73500 | - |

TABLE V. RESULT OF DATA AUGMENTATION AND DIFFERENT DROPOUT RATE OR NETWORK

| Exp_id | Backbone | Loss function | Dropout rate | Data aug | Remove bias | DB | Val_Acc | Test score |
|---|---|---|---|---|---|---|---|---|
| 9 | SeNet50 | BCE | 0.0 | no | no | no | 0.74625 | - |
| 10 | SeNet50 | BCE | 0.1 | no | no | no | 0.75625 | - |
| 11 | SeNet50 | Focal loss | 0.1 | no | no | no | 0.72500 | - |
| 12 | SeNet50 | BCE | 0.1 | yes | no | no | 0.76687 | - |
| 13 | SeNet50 | BCE | 0.1 | yes | yes | no | 0.76313 | - |
| 14 | SeNet50 | BCE | 0.1 | yes | no | yes | 0.77625 | 0.7078 |

TABLE VI. RESULT OF K-FOLD CROSS VALIDATION WITH DIFFERENT BACKBONE

| Exp_id | Backbone | Aug | K0 | K1 | K2 | K3 | K4 | K5 | K6 | mean | Test score |
|---|---|---|---|---|---|---|---|---|---|---|---|
| 15 | SeNet50 | Yes | 0.7275 | 0.7169 | 0.7438 | 0.7356 | 0.7094 | 0.7731 | 0.7719 | 0.7397 | 0.7336 |
| 16 | ResNet50 | Yes | 0.7262 | 0.7311 | 0.7525 | 0.7406 | 0.7006 | 0.7725 | 0.7744 | 0.74255 | 0.7422 |
| 17 | ResNet50 | No | 0.7244 | 0.7244 | 0.7338 | 0.7312 | 0.7069 | 0.7812 | 0.7775 | 0.73991 | - |

TABLE VII. RESULT OF K-FOLD CROSS VALIDATION WITH SAMPLE MORE NEGATIVE PAIRS, SYMMETRIC NETWORK STRUCTURE(EXP_ID 21), HIGH RESOLUTION INPUT(EXP_ID 20) AND MULTI-CLASS CLASSIFICATION METHOD(EXP_18)

| Exp_id | Pos:Neg | Size | K0 | K1 | K2 | K3 | K4 | K5 | K6 | mean | Test score |
|---|---|---|---|---|---|---|---|---|---|---|---|
| 18 | 1:2 | 128_197 | 0.7275 | 0.7404 | 0.7421 | 0.7442 | 0.7212 | 0.7483 | 0.7496 | 0.7390 | - |
| 19 | 1:2 | 128_197 | 0.7667 | 0.7608 | 0.7781 | 0.7735 | 0.7616 | 0.7884 | 0.7886 | 0.7739 | 0.7466 |
| 20 | 1:2 | 160_224 | 0.7667 | 0.7717 | 0.7946 | 0.7833 | 0.7717 | 0.8187 | 0.8058 | 0.7875 | 0.7510 |
| 21 | 1:2 | 128_197 | 0.7719 | 0.7679 | 0.7844 | 0.7781 | 0.7706 | 0.8087 | 0.8050 | 0.7838 | 0.7537 |

To reduce the distribution deviation between the predicted output of the softmax and the real label, using Cross-entropy loss function to calculate the uncertainty between the predicted output and the real label, where N is the class number.

$$L_{ce} = -\sum_{i=1}^{N} q_i \log(p_i) \begin{cases} q_i = 0, y \neq i \\ q_i = 1, y = i \end{cases} \quad (4)$$

III. EXPERIMENTS

In this section, we report our experimental results for the three tracks of the FG challenge.

A. Track 1

First, to verify the performance of the baseline and our adopted strategies (e.g., gray image, different backbone of VggFace, different loss functions), we randomly select 30 families as validation set and formulate the task as a binary classification problem. For mini-batch sizes, we randomly sample P/2 image pairs that have blood relative as positive samples and other P/2 image pairs that are not blood relative as negative samples. Finally, the batch size equals to B = P×2. In this experiment, we set P = 16 and resize image to 128x128 as FaceNet's input, and resize image to 197x197 as VggFace's input. By concatenating (x1+x2)^2, (x1-x2)^2, (x1-x2), (x1·x2), (x3 + x4)^2 etc., we get the results shown in Table IV.

From Table IV, we can see backbone ResNet50[11] and SeNet50[12] could get the same performance on validation set. BCE loss is better than Focal loss, and gray image shows inferior results. In particular, the baseline gets a score of 0.708 on the challenge stage.

In further experiments, we use backbone SeNet50 and randomly select 40 families as validation set. We also change the structure of the FC layers after concatenation (dropout rate, remove the bias of FC layers), apply data augmentation (random blurring, random change contrast and brightness, random flipping) and sample more image pairs from relationship 3,6,7,8(DB), as shown in Table V.

From Table V, we observe that increasing dropout rate, data augmentation, and DB could get better performance on validation set. Unfortunately, this does not yield a better score on the challenge stage.

In the next experiments, we split datasets into seven-folds for k-fold cross-validation, by setting dropout rate=0.1 with DB and data augmentation. We average every fold's result as the final result and get a score of 0.7336 with SeNet50 and a score of 0.7422 with ResNet50, as shown in Table VI.

Then we sample more negative pairs (exp_id: 19), try multi-class classification method (exp_id: 18), increase image size to 160x160 for FaceNet and 224x224 for VggFace (exp_id: 20) with ResNet50. Specifically, inspired by the importance of symmetric network structure, for example, (x1-x2) is not equal to (x2-x1), this will affect the performance of the model, so we remove $(x1^2-x2^2)$, $(x1-x2)$, $(x3^2-x4^2)$, $(x3-x4)$ and add $(x1–x2)^2$, $(x3-x4)^2$ (exp_id: 21) in the concatenate stage. Results are shown in Table VII.

From Table VII, we can see that sampling more negative pairs, higher resolution, and the symmetric network could achieve significant improvements. However, we are not allowed to conduct exhaustive tuning on our method due to the time limit. Finally, we average the result of experiment 16,20,21 and get our best score 0.7685, achieving 2nd place.

*B. Track 2*

The goal of task 2 is to decide whether a child is related to a pair of parents. Essentially, it is as same as task 1, and can be divided into two sub-problems of determining whether they are blood relatives or not between "Father-Child(FC)" and "Mother-Child(MC)".

In this task, firstly, we try to use triplet pairs "Father-Mother-Child" as input and randomly change child images to generate negative pairs and make binary predictions, but the performance is not as expected. To obtain better results, we employ a method similar to experiment 20 in task 1 and make predictions for FC and MC independently. Then we use two methods to get our final result. In the former approach, we use FC + MC as the final score and get score 0.7629 online. In the latter, we get results depend on FC and MC independently, then merge two results as the final result, achieving a score of 0.7578 online. We finally rank 3rd place in this task.

*C. Track 3*

The goal of task 3 is to find family members of the search subjects (i.e., the probes) in a search pool (i.e., the gallery), which can be seen as a many-to-many ranking problem.

In our method, we use the model of experiment 1 in task1 to get predictions of every "probe-gallery" pairs. After this, we could get a probability matrix for every probe member to all gallery members. Table VIII shows an example, with probability scores between every image of probe s0 and the entire gallery.

TABLE VIII. PROBABILITY MATRIX OF PROBES AND GALLERY

| Probe_s0 | Gallery | | | | |
|---|---|---|---|---|---|
| | 1 | 2 | 3 | 4 | … |
| 1 | $S_{11}$ | $S_{12}$ | $S_{13}$ | $S_{14}$ | … |
| 2 | $S_{21}$ | $S_{22}$ | $S_{23}$ | $S_{24}$ | … |
| 3 | $S_{31}$ | $S_{32}$ | $S_{33}$ | $S_{34}$ | … |
| … | … | … | … | … | … |
| Final score | $S_1$ | $S_2$ | $S_3$ | $S_4$ | … |

We explore different strategies of aggregation to obtain the final score (the last row of Table VIII), such as running average or the max value for each column. Finally, we achieving 4th place by averaging every column.

*D. Other*

We also try the deep metric learning-based method, by considering every person or person that are blood relatives as one class, using triplet loss[13] as loss function. We extract deep embedding features for every image, then measuring the similarity between image pairs via Euclidean distance. Despite its reasonable motivations, this does not outperform the previously described approach, which highlights further investigations.

IV. CONCLUSION

According to our experimental results, the method based on binary classification achieved our best score in all tasks and found that sampling more negative pairs, high resolution, and the symmetric network could bring significant improvements.

In the future, we will conduct more research on automatic kinship recognition, like face alignment, more efficient network structure, and analyze why deep metric learning-based method does not work in our experiments and so on.

*References*

[1] Joseph P. Robinson, Ming Shao, Yue Wu, Hongfu Liu, Timothy Gillis, Yun Fu, "Visual kinship recognition of families in the wild." In IEEE TPAMI, 2018.

[2] Joseph P. Robinson, Ming Shao, Handong Zhao, Yue Wu, Timothy Gillis, Yun Fu, "Recognizing Families In the Wild (RFIW): Data Challenge Workshop in conjunction with ACM MM 2017." In RFIW '17: Proceedings of the 2017 Workshop on Recognizing Families In the Wild, 2017.

[3] S. Wang, J. P. Robinson, and Y. Fu, "Kinship Verification on Families In The Wild with Marginalized Denoising Metric Learning." In 12th IEEE AMFG, 2017.

[4] Joseph P. Robinson, Ming Shao, Yue Wu, and Yun Fu, "Families in the Wild (FIW): Large-scale Kinship Image Database and Benchmarks." In Proceedings of the ACM on Multimedia Conference, 2016.

[5] Mattemilio, Northeastern SMILE Lab - Recognizing Faces in the Wild, Website, 2019.TEhttps://www.kaggle.com/c/recognizing-faces-in-the-wild/discussion/103670.

[6] Schroff F, Kalenichenko D, Philbin J. FaceNet: A unified embedding for face recognition and clustering[C]// 2015 IEEE Conference on Computer Vision and Pattern Recognition (CVPR). IEEE, 2015.

[7] Cao Q, Shen L, Xie W, et al. VGGFace2: A dataset for recognizing faces across pose and age[J]. 2017.

[8] Guo Y, Zhang L, Hu Y, et al. MS-Celeb-1M: A Dataset and Benchmark for Large-Scale Face Recognition[J]. 2016.

[9] Booth, David E . The Cross-Entropy Method[M]. Taylor & Francis Group, 2008.

[10] Lin T Y, Goyal P, Girshick R, et al. Focal Loss for Dense Object Detection[J]. IEEE Transactions on Pattern Analysis & Machine Intelligence, 2017, PP(99):2999-3007.

[11] He K, Zhang X, Ren S, et al. Deep Residual Learning for Image Recognition[C]// 2016 IEEE Conference on Computer Vision and Pattern Recognition (CVPR). IEEE Computer Society, 2016.

[12] Hu J, Shen L, Albanie S, et al. Squeeze-and-Excitation Networks[J]. IEEE Transactions on Pattern Analysis and Machine Intelligence, 2017.

[13] A. Hermans, L. Beyer, and B. Leibe. In defense of the triplet loss for person re-identification. arXiv preprint arXiv:1703.07737, 2017.